\documentclass[letterpaper, 10 pt, conference]{ieeeconf}  % Comment this line out if you need a4paper

\IEEEoverridecommandlockouts                              % This command is only needed if
\usepackage{graphicx}
\usepackage{subcaption}
\usepackage{multirow}
\let\labelindent\relax
\usepackage{enumitem}
\usepackage{amsmath}
\usepackage{booktabs}
\usepackage{amsfonts}
\usepackage{cite}
\usepackage{xcolor}
\usepackage{placeins}
\usepackage[table]{xcolor}
\title{\LARGE \bf
MatchFusion: Explicit–Implicit Instance Matching for Spatio-Temporal Multimodal Autonomous Driving
}

\author{
Xiaoyu Li$^{1,\dagger}$,
Jiajia Fu$^{1,\dagger}$,
Long Shi$^{1}$,
Tianyu Du$^{1}$,
Ruihang Li$^{2}$,\\[2pt]
Xian Wu$^{1}$,
Lijun Zhao$^{1,*}$,
Yingtao Zhang$^{1}$,
Lining Sun$^{1}$,
and Ruifeng Li$^{1}$%
\thanks{\raggedright
$^{1}$The authors are with Harbin Institute of Technology,
Harbin 150001, China.
E-mails:
{\tt\small
\{22S108236, 24s008051, 23s003065\}@stu.hit.edu.cn,
969491141@qq.com,
18929160507@163.com,
\{zhaolj, yingtao, Lnsnn, lrf100\}@hit.edu.cn}.}%
\thanks{\raggedright
$^{2}$Ruihang Li is with Zhejiang University,
Hangzhou 310027, China.
E-mail: {\tt\small 12221089@zju.edu.cn}.}%
\thanks{\raggedright
$^{\dagger}$Equal contribution.
$^{*}$Corresponding author.}%
}
\begin{document}

\maketitle
\thispagestyle{empty}
\pagestyle{empty}

%%%%%%%%%%%%%%%%%%%%%%%%%%%%%%%%%%%%%%%%%%%%%%%%%%%%%%%%%%%%%%%%%%%%%%%%%%%%%%%%

\begin{abstract}

Sparse instance representations provide a compact interface for spatial LiDAR–camera and temporal past–current interaction in multimodal perception and end-to-end autonomous driving (E2EAD).
Effective interaction requires reliable instance correspondences despite geometric discrepancies and heterogeneous semantic representations.
Attention-based methods exploit contextual semantics but often require specialized representation alignment, increasing computational overhead. 
In contrast, association based on structured object states is efficient and interpretable but lacks contextual evidence to resolve ambiguous matches.
To combine these complementary strengths, we propose MatchFusion, a learnable instance matching and fusion module for spatio-temporal multimodal autonomous driving.
MatchFusion initializes pairwise affinities using geometric similarity and category consistency, then selectively refines structurally plausible associations using instance embeddings.
The resulting soft matchmap guides a common residual aggregation operator for adaptive information exchange.
This unified matching--fusion formulation supports spatial LiDAR--camera and temporal past--current interaction, using multi-view image-plane geometry and motion-compensated Bird's-Eye-View (BEV) geometry as the respective structural priors.
Experiments on nuScenes demonstrate consistent perception gains across diverse front-end configurations.
Compared with a prior instance-centric fusion method, the MatchFusion-equipped system achieves higher perception accuracy while reducing FLOPs by 55.3\% and GPU memory usage by 39.3\%, with the matching–fusion module accounting for only 3.7\% of total perception latency.
Integrating temporal MatchFusion into SparseDrive further improves perception within an end-to-end driving framework without additional supervision.
These results establish explicit–implicit matching as an effective and efficient mechanism for spatio-temporal instance interaction.
Code will be released.

\end{abstract}

\section{INTRODUCTION}

Reliable multimodal autonomous driving benefits from integrating complementary observations across sensors~\cite{liao2025diffusiondrive, chitta2022transfuser} and time~\cite{sun2025sparsedrive, song2025momad}. 
LiDAR provides accurate spatial geometry, cameras capture rich appearance and semantics, and historical observations supply temporal context beyond individual frames.
Recent advances in sparse 3D perception~\cite{wang2025mv2dfusion,StreamCMT} and E2EAD~\cite{sun2025sparsedrive} increasingly represent scene entities as instance queries paired with structured object states.
These representations provide a compact interface for instance-level reasoning, reducing the need for dense interaction with background features.
Spatial LiDAR–camera and temporal past–current interaction can thus be formulated as information exchange between sparse instance sets.
Effective exchange requires reliable correspondences and efficient aggregation despite geometric discrepancies and heterogeneous semantic representations across modalities and time.

\begin{figure}[t]
    \centering
    \includegraphics[width=0.45\textwidth]{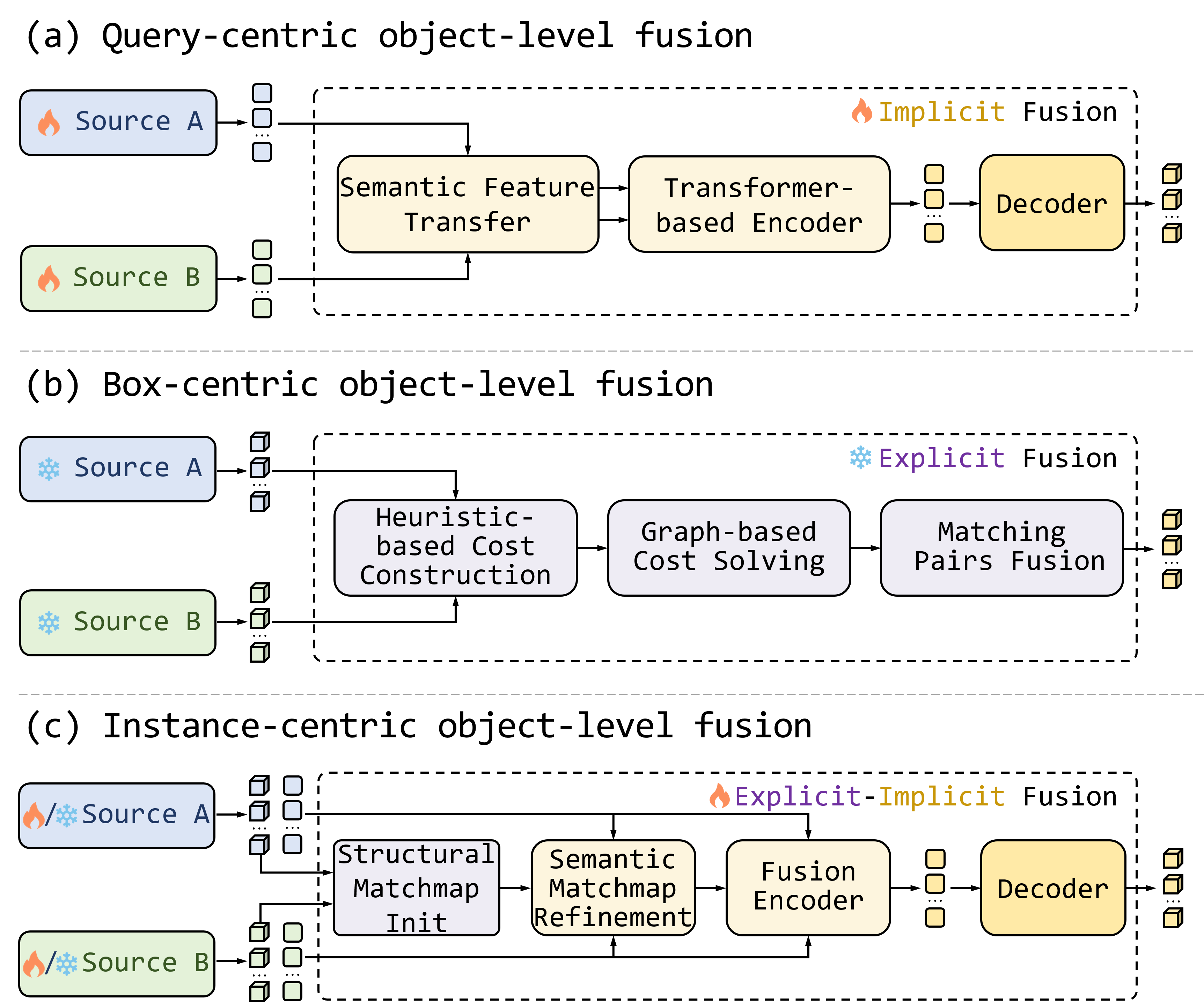}
    \caption{
    Various object-level multimodal fusion paradigms.
    }
    \label{fig:fusion_constrast}
\end{figure}

As illustrated in Fig.~\ref{fig:fusion_constrast}, existing instance-level interaction methods primarily adopt implicit attention-based interaction or explicit structural association.
Query-centric methods~\cite{xie2023sparsefusion, wang2025mv2dfusion} extract modality-specific instances and perform cross-modal interaction to integrate their features through learned transformations and attention.
These methods exploit rich contextual semantics and support adaptive information exchange, but aligning heterogeneous representations often requires specialized components that increase architectural complexity and computational overhead.
Temporal instance aggregation similarly relies on query propagation, feature compensation, and attention~\cite{lin2023sparse4d,wang2023exploring,sun2025sparsedrive} to accommodate object motion and viewpoint changes~\cite{li2026rethinking}.
In contrast, box-centric association methods~\cite{wu2026fusion,zhang2023bytetrackv2} construct costs from structural cues, such as geometry and category, and solve discrete assignments for state updates or candidate fusion.
Matching in a shared geometric space enables efficient and interpretable association without elaborate feature alignment.
However, hard assignments restrict fusion to selected correspondences, while structural costs alone cannot exploit contextual semantics to resolve ambiguous matches.
These complementary strengths motivate a matching and fusion mechanism that combines the efficiency of structural association with the adaptability of learned semantic interaction.

To this end, we propose MatchFusion, a learnable instance matching and fusion module for spatio-temporal multimodal autonomous driving. 
Given sparse instances comprising structured object states and query embeddings, MatchFusion initializes coarse correspondences by quantifying geometric similarity and category consistency. 
A lightweight semantic compatibility module then selectively refines structurally plausible associations using instance embeddings.
This coarse-to-fine process produces a soft matchmap that preserves many-to-many affinities and guides a common residual aggregation operator.
Spatial and temporal interaction instantiate this unified matching--fusion formulation with setting-specific geometric priors.
For LiDAR–camera interaction, 3D boxes from both modalities are projected onto image planes, and geometric similarities based on relative positions and sizes are aggregated across jointly visible views.
For past–current matching, past instances are aligned through object-motion compensation and ego-motion transformation, and geometric affinities are computed in BEV space.
Both settings use the same semantic refinement and residual aggregation mechanisms.
The aggregation operator is applied bidirectionally for cross-modal exchange and unidirectionally to incorporate past evidence into current candidates.
Temporal interaction can operate independently on upstream instances or incorporate spatially fused representations.

% For standalone detection, pretrained query-based front-ends remain frozen, enabling efficient training of the fusion module. 
Operating solely on instance representations, MatchFusion is decoupled from front-end feature extraction and provides enhanced queries for downstream perception and planning. 
Experiments on nuScenes~\cite{caesar2020nuscenes} demonstrate consistent perception gains across diverse front-end configurations. 
Compared with SparseFusion~\cite{xie2023sparsefusion}, the MatchFusion-equipped system achieves higher perception accuracy while reducing FLOPs by 55.3\% and GPU memory usage by 39.3\%, with the matching–fusion module accounting for only 3.7\% of total perception latency.
Integrating temporal MatchFusion into SparseDrive~\cite{sun2025sparsedrive} further improves perception within an end-to-end driving framework without additional supervision.
Our contributions are:

\begin{itemize}
\item We propose an explicit–implicit instance matching mechanism that combines structural affinity initialization with selective semantic refinement to establish efficient and adaptive soft correspondences.
\item We develop a unified matching--fusion formulation for spatial LiDAR--camera and temporal past--current interaction, combining setting-specific geometric priors with common semantic refinement and a single residual aggregation operator.
\item We demonstrate consistent perception gains across diverse front-ends, improved computational and memory efficiency, and compatibility with E2EAD frameworks.
\end{itemize}

\begin{figure*}[t]
    \centering
    \includegraphics[width=\textwidth]{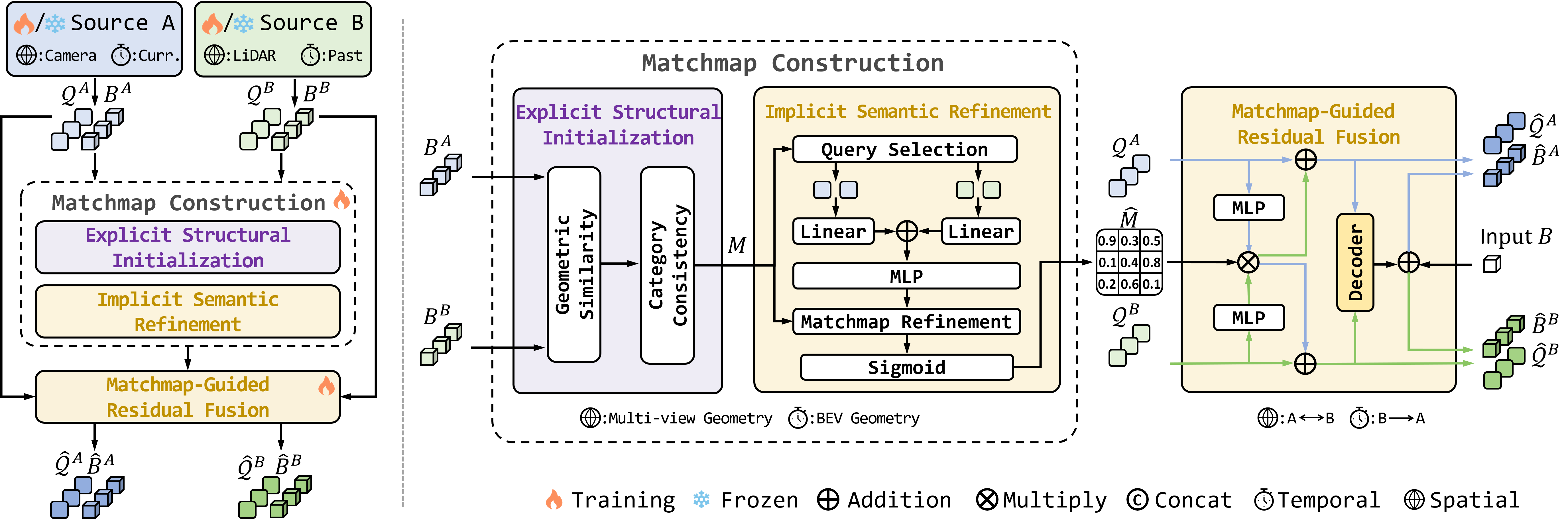}
    \caption{Architecture of MatchFusion for spatio-temporal interaction through unified instance matching and fusion.}
    \label{fig:framework}
    \vspace{-5mm}
    \end{figure*}

\section{RELATED WORK}

\textbf{Spatio-Temporal Multimodal 3D Perception.} 
Integrating complementary observations across sensors and time supports reliable scene understanding in autonomous driving.
For spatial fusion, LiDAR provides accurate geometry, while cameras capture rich appearance and semantics.
Point-level methods~\cite{vora2020pointpainting,wang2021pointaugmenting} augment point clouds with image-derived information.
BEVFusion~\cite{liu2023bevfusion} integrates modality-specific features in a shared BEV space.
SparseFusion~\cite{xie2023sparsefusion} and MV2DFusion~\cite{wang2025mv2dfusion} adopt sparse instance representations for object-level cross-modal interaction.
For temporal fusion, past information provides persistent context beyond individual frames. 
StreamPETR~\cite{wang2023exploring} propagates object queries with motion-aware feature compensation, Sparse4D~\cite{lin2023sparse4d} combines anchor-guided feature aggregation with recurrent instance propagation, and StreamCMT~\cite{StreamCMT} incorporates temporal query interaction into multimodal perception.
Spatio-temporal information further contributes to E2EAD by enriching the representations used for perception and planning.
LEAD~\cite{nguyen2026lead} highlights the value of multimodal information for driving.
BridgeAD~\cite{zhang2025bridging} uses sparse instance representations to connect temporal perception with motion prediction and planning.
These developments motivate a common interaction mechanism for integrating cross-modal and historical evidence. 
MatchFusion provides a unified matching–fusion formulation with geometric priors adapted to spatial LiDAR–camera and temporal past–current interaction.

\textbf{Instance Matching and Fusion.} 
Instance-level interaction requires establishing correspondences and aggregating complementary information.
Implicit attention-based approaches infer association weights from instance features and positional representations to guide feature aggregation.
SparseFusion~\cite{xie2023sparsefusion} combines modality-specific transformations with attention-based fusion. 
MV2DFusion~\cite{wang2025mv2dfusion} employs modality-specific query generators and uncertainty-aware positional representations for sparse interaction. 
Temporal methods similarly integrate past information through query propagation, feature compensation, and attention~\cite{wang2023exploring,sun2025sparsedrive}.
These methods exploit contextual semantics and support adaptive information exchange, but specialized representation processing can increase architectural complexity, computation, and memory consumption.
Explicit box-centric approaches~\cite{wu2026fusion,zhang2023bytetrackv2} instead construct association costs from geometric, categorical, or motion cues and solve discrete assignments for candidate fusion or state updates.
Matching compact structured states in a shared geometric space enables efficient and interpretable association without elaborate feature alignment.
However, hard assignments restrict information exchange to selected correspondences, and structural costs alone cannot exploit contextual evidence in instance embeddings to resolve ambiguous matches. 
MatchFusion combines the efficiency of structural association with the adaptability of semantic interaction through structural affinity initialization, selective semantic refinement, and soft matchmap-guided residual aggregation.

\section{MATCHFUSION}

\subsection{Overview and Instance Formulation}
\label{sec:overall}

As illustrated in Fig.~\ref{fig:framework}, MatchFusion is an instance-level matching and fusion module for spatio-temporal multimodal interaction.
It operates on sparse instance representations from upstream query-based perception models, decoupling matching and fusion from front-end feature extraction.
Each input source provides structured object states $\mathcal{B}\!=\!\{B_i\}_{i=1}^N$ and corresponding query embeddings $\mathcal{Q}\!=\!\{Q_i\}_{i=1}^N$, forming an instance set \(\mathcal{I}\!\!=\!\!\{I_i\!\!=\!\!(B_i, Q_i)\}_{i=1}^N\), where $N$ denotes the number of instances.
Each state $B_i$ comprises the 3D center $(x_i,y_i,z_i)$, dimensions $(w_i,l_i,h_i)$, yaw angle $\theta_i$, and predicted category $c_i$.
The query embedding $Q_i\in\mathbb{R}^{D}$ captures contextual semantics associated with the instance.

Given two instance sets, MatchFusion initializes coarse correspondences from structural cues and selectively refines them using query semantics. 
The resulting soft matchmap guides residual feature aggregation to produce enhanced instance representations for downstream perception and planning.
Sec.~\ref{sec:matchfusion} introduces this explicit–implicit matching and fusion mechanism. 
Sec.~\ref{sec:spatiotemporal} instantiates it for spatial LiDAR–camera and temporal past–current interaction.
MatchFusion separates instance interaction from task-specific prediction. 
Adapting it to spatial or temporal interaction requires specifying the geometric prior and aggregation direction, while the enhanced queries are optimized through the prediction objectives of the corresponding system.

\subsection{Explicit--Implicit Matching-Fusion}
\label{sec:matchfusion}

Structured object states provide compact and interpretable evidence for instance association, while query embeddings contain contextual information that helps resolve ambiguous candidates.
MatchFusion combines these complementary cues through a coarse-to-fine design: structural initialization establishes matching priors, and semantic refinement adjusts promising associations.
Given two instance sets, \(\mathcal{I}^A\!\!=\!\!\{I^A_i\!\!=\!\!(B^A_i, Q^A_i) \}_{i=1}^{N^A}\) and \(\mathcal{I}^B\!\!=\!\!\{I^B_j\!\!=\!\!(B^B_j, Q^B_j) \}_{j=1}^{N^B}\), the module performs explicit structural initialization, implicit semantic refinement, and matchmap-guided residual fusion.

\textbf{Explicit Structural Initialization.}
% Geometric and categorical cues provide an efficient starting point for association by comparing compact object states in a shared geometric space and enforcing category consistency.
Geometric similarity provides an efficient starting point for association by comparing compact object states in a shared geometric space. 
Category consistency is enforced as a hard constraint that excludes pairs with different predicted categories.
For each instance pair $I_i^A$ and $I_j^B$, the initial affinity is:
\begin{equation}
    M_{ij} = G_{ij} + H_{ij},
    \label{eq:matching similarity}
\end{equation}
where $G_{ij}$ represents geometric similarity and $H_{ij}$ is an additive category-consistency mask.
Geometric similarity is derived from relative object positions and sizes, with setting-specific formulations for spatial and temporal interaction detailed in Sec.~\ref{sec:spatiotemporal}.
Motivated by category-consistent association in prior 3D perception methods~\cite{li2023poly}, we define the mask as:
\begin{equation}
    H_{ij} = 
    \begin{cases}
        0, &  c^A_i = c^B_j, \\
        -\infty, & c^A_i\neq c^B_j,
    \end{cases}
\label{eq:category constraint}
\end{equation}
where $c_i^{A}$ and $c_j^{B}$ denote the predicted categories.
Category-consistent pairs retain their geometric affinities, whereas category-inconsistent pairs are excluded from subsequent semantic refinement and matchmap-guided aggregation.
These cues yield the initial affinity matrix \(M\in\mathbb{R}^{N^{A}\times N^{B}}\) without elaborate transformations of query embeddings.
% Since geometry and category cues naturally share a low-context space, they provide an efficient prior for subsequent semantic refinement without additional alignment.

\textbf{Implicit Semantic Refinement.}
Structural similarity can leave ambiguities among geometrically similar candidates. 
Query embeddings provide contextual evidence to refine these associations.
We condition semantic refinement on the initial masked affinity, concentrating learned pairwise corrections on category-consistent pairs.
\begin{equation}
\tilde{M}_{ij}=
\begin{cases}
\quad M_{ij},\quad M_{ij} \leq \epsilon, \\ 
M_{ij}+{R}_{ij},\quad M_{ij}> \epsilon,
\end{cases} \\    
\end{equation}
Here, \(\epsilon\) is a finite threshold controlling semantic refinement. 
Category-inconsistent pairs receive no semantic correction.
% For category-consistent pairs, the selection criterion reduces to \(G_{ij}>\epsilon\).
For selected pairs, the semantic correction is computed as:
\begin{equation}
R_{ij} =
\mathrm{MLP}\!\left(
\psi_A(Q_i^{A}) + \psi_B(Q_j^{B})
\right),
\label{eq:semantic_correction}
\end{equation}
where $\psi_A$ and $\psi_B$ are branch-specific linear mappings.
The multi-layer perceptron (MLP) maps their summed embeddings to a scalar correction.
An element-wise sigmoid function converts the resulting affinities into matching scores:
\begin{equation}
     \hat{M}_{ij}=\mathrm{Sigmoid}(\tilde{M}_{ij}).
\end{equation}

% Pairs at or below the refinement threshold receive no semantic correction but retain their sigmoid-transformed structural affinities.
The resulting soft matchmap $\hat{M}$ preserves graded many-to-many associations, allowing contextual semantics to refine promising correspondences without imposing discrete assignments~\cite{kim2021eagermot}.

\textbf{Matchmap-Guided Residual Fusion.}
Ambiguous observations can support multiple plausible correspondences, making early discrete assignment restrictive.
% \textcolor{red}{The soft matchmap $\hat{M}$ instead guides many-to-many feature aggregation, incorporating complementary evidence through residual feature updates.}
The soft matchmap $\hat{M}$ instead guides many-to-many feature aggregation among category-consistent instances, incorporating complementary evidence through residual feature updates.
For information flow from $\mathcal{I}^{B}$ to $\mathcal{I}^{A}$, we define a common residual aggregation operator:
\begin{equation}
\mathcal{F}(Q^{A}, Q^{B}, \hat{M})
=
Q^{A}
+
\hat{M}\,\otimes\,\mathrm{MLP}_{B\rightarrow A}(Q^{B}),
\label{eq:residual_fusion}
\end{equation}
where $\otimes$ denotes matrix multiplication and
$\mathrm{MLP}_{B\rightarrow A}$ maps source embeddings into the receiving feature space.

The same operator supports both spatial and temporal fusion.
Spatial interaction applies $\mathcal{F}$ in both directions using
$\hat{M}$ and $\hat{M}^{\mathsf{T}}$, with both updates operating on the original input queries.
Temporal interaction applies $\mathcal{F}$ once to aggregate aligned historical features into augmented current candidates.
The enhanced queries are subsequently processed by the corresponding prediction heads, separating the common aggregation mechanism from task-specific prediction.

\subsection{Spatio-Temporal Instantiation}
\label{sec:spatiotemporal}

MatchFusion supports spatio-temporal interaction through a common explicit–implicit matching and residual aggregation principle. 
Setting-specific geometric priors and candidate selection establish plausible correspondences, while the aggregation direction determines how complementary evidence updates instance representations.

\textbf{Spatial LiDAR--Camera Interaction.}
Camera depth errors can separate corresponding instances in 3D while preserving their projected overlap in relevant views~\cite{philion2020lift}.
We therefore initialize cross-modal affinities using multi-view image-plane geometry to reduce sensitivity to depth discrepancies. 
Since distinct objects can also exhibit similar projections, these affinities provide coarse priors for subsequent semantic refinement.
For each camera $k=1,\ldots,N_{\text{view}}$, we project 3D boxes from both modalities onto the image plane:
\begin{equation}
    b^{C,k}_i \leftarrow \mathrm{Proj}\left( B^{C}_i, P^k \right),b^{L,k}_j \leftarrow \mathrm{Proj}\left( B^{L}_j, P^k \right),
\end{equation}
where $\mathrm{Proj}(\cdot)$ denotes 3D-to-2D projection and $P^k$ is the corresponding projection matrix.
Each projected box is represented by its center and extent, $b=(bx,by,bw,bh)$.
Relative geometry is encoded using normalized positional offsets and logarithmic size ratios:
{
\begin{multline}
g_{ij}^k=
\Big[\log\!\left(\tfrac{|bx^{C, k}_i-bx^{L, k}_j|}{bw^{C, k}_i}{+}1\right),
\log\!\left(\tfrac{|by^{C, k}_i-by^{L, k}_j|}{bh^{C, k}_i}{+}1\right), \\
\log\!\left(\tfrac{bw^{C, k}_i}{bw^{L, k}_j}\right),
\log\!\left(\tfrac{bh^{C, k}_i}{bh^{L, k}_j}\right)\Big].
\label{eq:geomtric similarity}
\end{multline}
}

Sinusoidal positional encoding followed by an MLP maps this descriptor to a view-specific geometric affinity:
% \begin{equation}
% \hat{g}_{ij}^k = \mathrm{PosEmb}(g_{ij}^k),
% \end{equation}
\begin{equation}
G_{ij}^k = \mathrm{MLP}\big(\mathrm{Flatten}(\mathrm{PosEmb}(g_{ij}^k))\big),
\end{equation}
where $\mathrm{PosEmb}(\cdot)$ denotes sinusoidal positional encoding.
For pairs with at least one jointly visible view, we aggregate affinities through visibility-aware averaging:
\begin{equation}
G_{ij} = \frac{\sum_{k=1}^{N_{\text{view}}} O_{ij}^k \cdot G_{ij}^k}{\sum_{k=1}^{N_{\text{view}}} O_{ij}^k},
\end{equation}
where $O_{ij}^k=1$ if both projected boxes are visible in view $k$, and zero otherwise.
The aggregated affinity is combined with category consistency through Eq.~\ref{eq:matching similarity} and refined using query semantics as described in Sec.~\ref{sec:matchfusion}.
The resulting matchmap $\hat{M}$ guides bidirectional application of the residual aggregation operator in Eq.~\ref{eq:residual_fusion}:
\begin{equation}
\begin{aligned}
\hat{Q}^{C} &= \mathcal{F}(Q^{C},Q^{L},\hat{M}),\\
\hat{Q}^{L} &= \mathcal{F}(Q^{L},Q^{C},\hat{M}^{\mathsf{T}}).
\end{aligned}
\end{equation}

Both updates use the original input queries.
The enhanced queries are concatenated along the instance dimension and passed to an added refinement decoder. 
Its regression branch predicts box offsets:
\begin{equation}
\Delta\mathcal{B}
=
\mathrm{Decoder}
\left(
\mathrm{Concat}
\left(
\hat{\mathcal{Q}}^{C},
\hat{\mathcal{Q}}^{L}
\right)
\right),
\end{equation}
which update the corresponding input boxes:
\begin{equation}
\hat{\mathcal{B}}
=
\mathrm{Concat}
\left(
\mathcal{B}^{C},
\mathcal{B}^{L}
\right)
+
\Delta\mathcal{B}.
\end{equation}

Retaining candidates from both modalities preserves complementary object hypotheses and semantic evidence.
The resulting queries and updated states form the spatially fused instance set \(\hat{\mathcal{I}}\).
Classification and box-regression objectives supervise the added decoder.

\textbf{Temporal Past--Current Interaction.}
Historical instances provide persistent object information under weak or incomplete current observations.
Object-motion compensation and ego-motion transformation align historical states $\mathcal{I}^{t-1}$ to the current frame, yielding $\mathcal{I}^{t-1,t}$.
These aligned instances are combined with the current set $\mathcal{I}^{t}$ to form an augmented candidate set $\bar{\mathcal{I}}$, preserving historical hypotheses within the available query budget.
Notably, current instances can originate from an upstream perception branch or from spatially fused set $\hat{\mathcal{I}}$, allowing temporal interaction to operate independently or alongside spatial fusion.

Temporal matching associates augmented current candidates $\bar{\mathcal{I}}$ with aligned historical instances $\mathcal{I}^{t-1,t}$.
After motion compensation, relative BEV positions and sizes provide geometric priors for association.
We therefore construct geometric affinities directly in the BEV space, while semantic refinement helps resolve ambiguities caused by residual localization and motion estimation errors.
For augmented candidate $\bar{I}_i$ and aligned historical instance $I_j^{t-1,t}$, we define:
\begin{multline}
g_{ij}=
\Big[\log\!\left(\tfrac{|\bar{x}_i-x^{t-1,t}_j|}{\bar{w}_i}{+}1\right),
\log\!\left(\tfrac{|\bar{y}_i-y^{t-1,t}_j|}{\bar{l}_i}{+}1\right),\\
\log\!\left(\tfrac{\bar{w}_i}{w^{t-1,t}_j}\right),
\log\!\left(\tfrac{\bar{l}_i}{l^{t-1,t}_j}\right)\Big].
\label{eq:temporal_geometry}
\end{multline}

As in spatial interaction, sinusoidal positional encoding followed by an MLP maps the descriptor to a scalar affinity:
\begin{equation}
G_{ij} = \mathrm{MLP}\big(\mathrm{Flatten}(\mathrm{PosEmb}(g_{ij}))).
\end{equation}

Category consistency and semantic refinement then produce the temporal matchmap $\bar{M}$.
The residual aggregation operator incorporates historical evidence into current queries:
\begin{equation}
\bar{Q}'
=
\mathcal{F}(\bar{Q},Q^{t-1,t},\bar{M}).
\end{equation}

This unidirectional update enriches current candidates with past context.
Enhanced queries $\bar{Q}'$ continue through the host decoder to refine the associated object states, yielding $\bar{\mathcal{I}}'$.

\section{EXPERIMENTS}
\label{sec:experiments}

\subsection{Datasets and Implementation Details}
\label{sec:experimental_setup}

\textbf{Datasets and Metrics.}
% We evaluate MatchFusion on nuScenes~\cite{caesar2020nuscenes} for perception and open-loop planning, and on NeuroNCAP~\cite{ljungbergh2024neuroncap} for closed-loop driving.
% For nuScenes \textbf{detection}, we report the nuScenes detection score (NDS) and mean average precision (mAP).
% For \textbf{tracking}, we report average multi-object tracking accuracy (AMOTA) and multi-object tracking accuracy (MOTA).
% For \textbf{open-loop planning} on nuScenes, we evaluate trajectory L2 error and Collision Rate (CR).
% For \textbf{closed-loop planning} on NeuroNCAP, we report the NeuroNCAP Score and closed-loop Collision Rate.
% Computational efficiency is evaluated using FLOPs, GPU memory, and inference latency.
We evaluate MatchFusion on nuScenes for perception and planning. 
For nuScenes \textbf{detection}, we report the nuScenes detection score (NDS) and mean average precision (mAP). 
\textbf{Tracking} is evaluated using average multi-object tracking accuracy (AMOTA) and multi-object tracking accuracy (MOTA). 
\textbf{Planning} is assessed using trajectory L2 error and collision rate (CR). 
Computational efficiency is measured by FLOPs, GPU memory consumption, and inference latency.

\textbf{\textcolor{black}{Implementation Details.}}
We denote the spatial, temporal, and combined spatio-temporal configurations as MatchFusion-S, MatchFusion-T, and MatchFusion-ST, respectively. 
MatchFusion-S enables LiDAR–camera interaction, while MatchFusion-T aggregates historical evidence into current instance representations.
Category consistency is enforced across all settings.
Each component of the 4-D relative geometric descriptor is encoded using a 256-D sinusoidal embedding, yielding a flattened 1024-D feature that a single-layer MLP maps to geometric affinity. 
Semantic refinement uses a single-layer MLP. 
The residual-fusion and modality-projection MLPs each comprise two layers. 
All hidden representations are 256-D.
For perception experiments, we freeze the pretrained front-ends to isolate the contribution of MatchFusion and evaluate its ability to enhance existing instance representations without updating the front-ends. 
We evaluate multiple front-end configurations and train MatchFusion for 16 epochs with a batch size of 32 on four NVIDIA RTX 3090 GPUs, requiring approximately 12 hours.
We use AdamW~\cite{adamw} with an initial learning rate of $1.0\times10^{-3}$ and cosine annealing schedule. 
All perception results are reported on the nuScenes validation set without test-time augmentation or model ensembling.
In MatchFusion-S, enhanced camera and LiDAR queries are concatenated and processed by an added refinement decoder, supervised using the classification and box-regression objectives adopted by Sparse4Dv3~\cite{lin2023sparse4d}. 
MatchFusion-T uses existing downstream objectives without additional losses. 
Tracking follows the query propagation and update mechanism of Sparse4Dv3 without explicit association module.
For E2EAD, we integrate MatchFusion-T into SparseDrive and jointly train the entire system from scratch, following the baseline training protocol. 
This setting evaluates its compatibility with end-to-end optimization under the original perception, motion, and planning objectives.
MatchFusion-T operates directly on the instance representations of framework without spatial LiDAR–camera fusion or additional supervision. 
The baseline and MatchFusion-equipped systems follow matched training and evaluation protocols.

% As shown in Table.~\ref{tab:frontend_generalization}, we evaluate whether MatchFusion consistently improves perception across different upstream instance representations. 
% With TransFusion-L~\cite{bai2022transfusion} as the LiDAR front-end, we pair MatchFusion with four camera detectors: DETR3D~\cite{DETR3D}, StreamPETR~\cite{wang2023exploring}, Sparse4Dv3~\cite{lin2023sparse4d}, and SimPB~\cite{tang2025simpb}. 
% Across these configurations, MatchFusion improves over the stronger unimodal front-end by 1.9–3.1 NDS points and 4.3–5.5 mAP points. 
% Even with the comparatively weak DETR3D camera detector, fusion increases NDS from 70.1 to 72.0 and mAP from 65.1 to 69.4, showing that complementary camera evidence remains useful despite a substantial accuracy gap between the two front-ends. 
% These consistent gains support the compatibility of the matching–fusion interface with different sparse instance representations.
% \textcolor{red}{For the Sparse4Dv3-based configuration, the expected improvements in AMOTA and AMOTP would further demonstrate that enhanced instance representations benefit tracking alongside detection. 
% Since tracking follows the existing query propagation and update mechanism, these gains would indicate that MatchFusion improves the representations used by the downstream tracker without requiring an additional explicit association module.}

\subsection{Comparative Evaluation}
\label{sec:comparison}

\textbf{Perception Gains Across Front-Ends.}
The instance-level formulation of MatchFusion separates matching and fusion from front-end feature extraction. 
To validate the compatibility of MatchFusion with diverse instance representations, we retain TransFusion-L as LiDAR front-end and pair it with DETR3D, StreamPETR, Sparse4Dv3, and SimPB.
As shown in Table.~\ref{tab:frontend_generalization}, MatchFusion-ST consistently improves over the stronger unimodal front-end, yielding gains of 1.9–3.1\% NDS and 4.3–5.5\% mAP. 
With the weak DETR3D camera detector, fusion improves NDS from 70.1 to 72.0 and mAP from 65.1 to 69.4.
This demonstrates that a less accurate front-end can still provide complementary evidence through effective instance matching and aggregation.
These consistent improvements support the reuse of MatchFusion across different camera architectures through a common instance interface.
% tracking task
For the Sparse4Dv3-based configuration, MatchFusion-ST also improves AMOTA and MOTA through the existing query propagation mechanism, extending its benefits from detection to tracking.

\textbf{Accuracy--Efficiency Trade-off.}
To demonstrate the efficiency of MatchFusion for integrated perception, we evaluate system-level computational cost and module-level inference latency.
With Sparse4Dv3 and TransFusion-L as front-ends, MatchFusion-ST-equipped system achieves 73.0 NDS and 62.3 AMOTA with 254.2 GFLOPs and 3.7 GB memory. 
As shown in Table~\ref{tab:accuracy_efficiency}, it achieves higher NDS than SparseFusion while reducing FLOPs by 55.3\% and memory consumption by 39.3\%.
These results demonstrate that combining compact structural priors with lightweight semantic refinement enables effective instance interaction without introducing substantial computational overhead.
Furthermore, under the same inference setting, the camera front-end, LiDAR front-end, and MatchFusion-ST require 47.3, 283.3, and 12.9 ms, respectively.
Including tracking operations, MatchFusion-ST accounts for only 3.7\% of the total latency, supporting integrated detection and tracking with limited additional runtime.
This further demonstrates that the proposed instance interaction mechanism enables accurate integrated perception with only marginal runtime overhead.

\begin{table*}[t]
  \centering
  \caption{
  Comparison on nuScenes for the E2EAD task.
  $\dagger$: Our reproduced results by official code.
  }
  \label{tab:e2e_comparison}
  \setlength{\tabcolsep}{4pt}
  \renewcommand{\arraystretch}{0.95}
  \small
  \resizebox{\textwidth}{!}{%
  \begin{tabular}{c|c|cc|cc|cc}
    \toprule
    \multirow{2}{*}[-0.5ex]{Dataset}
    & \multirow{2}{*}[-0.5ex]{Method}
    & \multicolumn{2}{c|}{Detection}
    & \multicolumn{2}{c|}{Tracking}
    & \multicolumn{2}{c}{Planning} \\

    \cmidrule(lr){3-4}
    \cmidrule(lr){5-6}
    \cmidrule(lr){7-8}

    &
    & NDS$\uparrow$
    & mAP$\uparrow$
    & AMOTA$\uparrow$
    & MOTA$\uparrow$
    & L2 (m)$\downarrow$
    & CR (\%)$\downarrow$ \\
    % & DS$\uparrow$
    % & SR (\%)$\uparrow$ \\
    \midrule

    \multirow{2}{*}{nuScenes}
    & SparseDrive$\dagger$
    & 52.2
    & 41.2
    & 36.9
    & 34.2
    & \textbf{0.63}
    & 0.123 \\

    & SparseDrive + MatchFusion-T
    & \textbf{53.0 (+0.8)}
    & \textbf{41.7 (+0.5)}
    & \textbf{39.6 (+2.7)}
    & \textbf{35.9 (+1.7)}
    & \textcolor{black}{0.64}
    & \textcolor{black}{\textbf{0.119}} \\

    % \midrule

    % \multirow{2}{*}{\textcolor{red}{NeuroNCAP}}
    % & SparseDrive
    % & --
    % & --
    % & --
    % & --
    % & --
    % & -- \\

    % & SparseDrive + MatchFusion-T
    % & --
    % & --
    % & --
    % & --
    % & --
    % & -- \\

    \bottomrule
  \end{tabular}
  }
\end{table*}

\begin{table}[t]
  \centering
  \caption{
  Perception enhancement with MatchFusion-ST across front-end configurations on the nuScenes validation set.
  LiDAR denotes the TransFusion-L front-end.
  }
  \label{tab:frontend_generalization}
  \setlength{\tabcolsep}{5pt}
  \begin{tabular}{
    @{}
    c
    @{\hspace{0.5pt}}
    c
    c|
    c @{\hspace{1pt}} c
    @{}
  }
    \toprule
    Camera
    & LiDAR
    & MatchFusion-ST
    & NDS$\uparrow$
    & mAP$\uparrow$ \\
    \midrule

    --
    & \checkmark
    & --
    & 70.1
    & 65.1 \\[2pt]

    \multirow{2}{*}{DETR3D~\cite{DETR3D}}
    & --
    & --
    & 42.2
    & 34.7 \\
    & \checkmark
    & \checkmark
    & 72.0 \textbf{(+1.9)}
    & 69.4 \textbf{(+4.3)} \\[2pt]

    \multirow{2}{*}{StreamPETR~\cite{wang2023exploring}}
    & --
    & --
    & 53.7
    & 43.2 \\
    & \checkmark
    & \checkmark
    & 72.7 \textbf{(+2.6)}
    & 69.9 \textbf{(+4.8)} \\[2pt]

    \multirow{2}{*}{Sparse4Dv3~\cite{lin2023sparse4d}}
    & --
    & --
    & 56.1
    & 46.9 \\
    & \checkmark
    & \checkmark
    & 73.0 \textbf{(+2.9)}
    & 70.4 \textbf{(+5.3)} \\[2pt]

    \multirow{2}{*}{SimPB~\cite{tang2025simpb}}
    & --
    & --
    & 59.0
    & 48.7 \\
    & \checkmark
    & \checkmark
    & \textbf{73.2} \textbf{(+3.1)}
    & \textbf{70.6} \textbf{(+5.5)} \\
    \bottomrule
  \end{tabular}
\end{table}

\begin{table}[t]
  \centering
  \caption{
  Detection accuracy and computational cost on the nuScenes validation set.
  All methods use VoxelNet~\cite{zhou2018voxelnet} and ResNet-50~\cite{he2016deep} as the LiDAR and camera backbones, respectively.
  FLOPs and GPU memory are measured for the complete system with a batch size of one.
  $\dagger$: Our reproduced results by official code.
  }
  \label{tab:accuracy_efficiency}
  \setlength{\tabcolsep}{0pt}
  \renewcommand{\arraystretch}{1.1}
  \begin{tabular*}{\columnwidth}{
    @{}
    c
    @{\hspace{3pt}}|
    @{\hspace{3pt}\extracolsep{\fill}}
    c
    @{}
    c
    @{\hspace{2pt}}
    c
    @{\hspace{0pt}}
    c
    @{}
  }
    \toprule
    \multirow{2}{*}{Method}
    & \multicolumn{2}{c}{Accuracy}
    & \multicolumn{2}{c@{}}{Efficiency} \\
    \cmidrule(lr){2-3}
    \cmidrule(l){4-5}
    & NDS$\uparrow$
    & mAP$\uparrow$
    & FLOPs (G)$\downarrow$
    & Mem. (GB)$\downarrow$ \\
    \midrule
    TransFusion$\dagger$~\cite{bai2022transfusion}
    & 71.3 & 67.5
    & 449.8 & 10.5 \\
    DeepInteraction$\dagger$~\cite{yang2022deepinteraction}
    & 72.6 & 69.9
    & 513.1 & 22.1 \\
    SparseFusion~\cite{xie2023sparsefusion}
    & 72.8 & \textbf{70.4}
    & 569.1 & 6.1 \\
    DeepInteraction++~\cite{yang2025deepinteraction++}
    & 72.9 & 70.1
    & -- & 11.4 \\
    StreamCMT~\cite{StreamCMT}
    & 71.8 & 69.0
    & -- & -- \\
    \addlinespace[1pt]
    \midrule
    \addlinespace[1pt]
    MatchFusion-ST (Ours)
    & \textbf{73.0}
    & \textbf{70.4}
    & \textbf{254.2}
    & \textbf{3.7} \\
    \bottomrule
  \end{tabular*}
  \vspace{-3mm}
\end{table}

\begin{table}[t]
  \centering
  \caption{
  Contributions of spatial and temporal interaction.
  }
  \label{tab:spatiotemporal_ablation}

  \setlength{\tabcolsep}{5pt}

  \begin{tabular}{@{}cc c| cc cc@{}}
    \toprule
    \multicolumn{2}{c}{Spatial}
    & \multirow{2}{*}{Temporal}
    & \multicolumn{2}{c}{Detection}
    & \multicolumn{2}{c}{Tracking} \\
    \cmidrule(lr){1-2}
    \cmidrule(lr){4-5}
    \cmidrule(lr){6-7}
    Camera & LiDAR &
    & NDS$\uparrow$ & mAP$\uparrow$
    & AMOTA$\uparrow$ & MOTA$\uparrow$ \\
    \midrule
    -- & \checkmark & --
    & 70.1 & 65.1
    & -- & -- \\
    \checkmark & -- & --
    & 56.1 & 46.9
    & 49.0 & 43.6 \\
    \checkmark & \checkmark & --
    & 72.5 & 69.6
    & 49.0 & 43.6 \\
    \checkmark & \checkmark & \checkmark
    & \textbf{73.0} & \textbf{70.4}
    & \textbf{62.3} & \textbf{60.8} \\
    \bottomrule
  \end{tabular}
\end{table}

% \begin{table}[t]
%   \centering
%   \caption{
%   Detection accuracy and computational cost on the nuScenes validation set.
%   All methods use VoxelNet~\cite{zhou2018voxelnet} and ResNet-50~\cite{he2016deep} as the LiDAR and camera backbones, respectively.
%   FLOPs and GPU memory are measured for the complete system with a batch size of one.
%   }
%   \label{tab:accuracy_efficiency}

%   \setlength{\tabcolsep}{0.5pt}
%   \renewcommand{\arraystretch}{1.1}

%   \begin{tabular*}{\columnwidth}{
%     @{\extracolsep{\fill}}
%     l
%     cc
%     cc
%     @{}
%   }
%     \toprule
%     \multirow{2}{*}{\multicolumn{1}{c}{Method}}
%     & \multicolumn{2}{c}{Accuracy}
%     & \multicolumn{2}{c}{Efficiency} \\
%     \cmidrule(lr){2-3}
%     \cmidrule(lr){4-5}
%     & NDS$\uparrow$
%     & mAP$\uparrow$
%     & FLOPs (G)$\downarrow$
%     & Mem. (GB)$\downarrow$ \\
%     \midrule

%     TransFusion~\cite{bai2022transfusion}
%     & 71.3 & 67.5
%     & 449.8 & 10.5 \\

%     DeepInteraction~\cite{yang2022deepinteraction}
%     & 72.6 & 69.9
%     & 513.1 & 22.1 \\

%     SparseFusion~\cite{xie2023sparsefusion}
%     & 72.8 & \textbf{70.4}
%     & 569.1 & 6.1 \\

%     \textcolor{red}{NEWMETHOD}
%     & XX & XX
%     & XX & XX \\

%     \addlinespace[1pt]
%     \midrule
%     \addlinespace[1pt]

%     MatchFusion (Ours)
%     & \textbf{73.0}
%     & \textbf{70.4}
%     & \textbf{254.2}
%     & \textbf{\textcolor{red}{4.1}} \\

%     \bottomrule
%   \end{tabular*}
% \end{table}

\begin{table}[t]
  \centering
  \caption{
  Ablation of matchmap construction and fusion strategies on MatchFusion-S.
  }
  \label{tab:matching_fusion_ablation}
  \label{tab:matchmap_ablation}
  \label{tab:fusion_strategy}
  \setlength{\tabcolsep}{4pt}
  \renewcommand{\arraystretch}{0.9}
  \begin{tabular}{@{}cc|cc@{}}
    \toprule
    \multirow{2}{*}{Setting} & \multirow{2}{*}{Variant}
    & \multicolumn{2}{c}{Detection} \\
    \cmidrule(lr){3-4}
    &
    & NDS$\uparrow$ & mAP$\uparrow$ \\
    \midrule
    \rowcolor{gray!10}
    \multicolumn{4}{c}{\textit{Matching}} \\
    \multirow{2}{*}{3D}
      & Structural & 72.2 & 69.1 \\
      & Structural + Semantic & 72.4 & 69.2 \\[2pt]
    2D
      & Structural + Semantic & \textbf{72.5} & \textbf{69.6} \\
    \midrule
    \rowcolor{gray!10}
    \multicolumn{4}{c}{\textit{Fusion}} \\
    \multirow{3}{*}{Implicit}
      & Concat & 69.4 & 64.8 \\
      & Self-Attention & 71.3 & 67.3 \\
      & Cross-Attention & 71.8 & 67.9 \\[2pt]
    Explicit--Implicit
      & MatchFusion-S & \textbf{72.5} & \textbf{69.6} \\
    \bottomrule
  \end{tabular}
  
\end{table}

\begin{table}[t]
  \centering
  \caption{Ablation of the category consistency mechanism in MatchFusion-ST.}
  \label{tab:category_consistency_ablation}
  \setlength{\tabcolsep}{12pt}
  \renewcommand{\arraystretch}{0.9}
  \begin{tabular}{@{}ccc@{}}
    \toprule
    Category Consistency & NDS$\uparrow$ & mAP$\uparrow$ \\
    \midrule
    $\times$    & 72.1 & 69.0 \\
    \checkmark  & \textbf{73.0} & \textbf{70.4} \\
    \bottomrule
  \end{tabular}
\end{table}

\begin{table}[t]
  \centering
  \caption{Training efficiency of MatchFusion-S matchmap variants in the 2D setting. 
  Time and GPU memory are normalized to the structural baseline.}
  \setlength{\tabcolsep}{7pt}
  \begin{center}
    \begin{tabular}{c|cc}
    \toprule
    Matchmap Construction
    & Mem.$\downarrow$
    & Time$\downarrow$ \\
    \midrule
    Structural
    & 1.0$\times$
    & 1.0$\times$ \\
    Structural + Semantic
    & 1.3$\times$
    & 1.7$\times$ \\
    Semantic
    & 2.2$\times$
    & 1.9$\times$ \\
    \bottomrule
\end{tabular}
  \end{center}
  \label{tab:matchmap_efficiency}
  \vspace{-5mm}
\end{table}

\textbf{Integration into End-to-End Driving.}
% To evaluate downstream utility and the independent applicability of temporal interaction, we integrate MatchFusion-T into SparseDrive without spatial LiDAR–camera fusion.
% The enhanced queries are optimized through the original perception, motion, and planning objectives.
% As shown in Table~\ref{tab:e2e_comparison}, MatchFusion-T improves NDS and mAP \textcolor{black}{by 0.8 and 0.5 points}, respectively, and AMOTA and MOTA by \textcolor{black}{2.7 and 1.7 points}. 
% Planning CR decreases from \textcolor{black}{0.123\% to 0.119\%}.
% These consistent gains indicate that temporal matching enriches the shared instance representations used across tasks.
% % The improvement in planning further supports the representation-level formulation of MatchFusion, showing that its benefits extend beyond object-state refinement to the features directly involved in driving decisions.
% % On NeuroNCAP, the improvements in DS and SR provide complementary evidence under closed-loop execution, where planning decisions directly affect subsequent observations.
% % \textcolor{red}{On NeuroNCAP, the higher NeuroNCAP Score and lower collision rate provide complementary evidence under closed-loop execution, where planning decisions directly affect subsequent observations.}
% Consistent gains demonstrate that the proposed matching--fusion principle transfers effectively from integrated perception to end-to-end autonomous driving.
To evaluate the independent applicability of temporal interaction within an end-to-end driving framework, we integrate MatchFusion-T into SparseDrive using its original training objectives without additional supervision. 
As shown in Table I, MatchFusion-T improves NDS and mAP by 0.8 and 0.5 points, respectively, and AMOTA and MOTA by 2.7 and 1.7 points. 
Planning performance remains comparable, with CR decreasing from 0.123\% to 0.119\% and L2 error increasing slightly from 0.63 to 0.64 m. 
These results support the applicability of temporal MatchFusion to end-to-end driving, with the clearest gains observed in perception.

\subsection{Ablation Studies}
\label{sec:ablation}

Unless otherwise specified, ablation studies use Sparse4Dv3 with ResNet-50 and TransFusion-L with VoxelNet as the camera and LiDAR front-ends, respectively. 
We examine the contributions of spatial and temporal interaction, the design of the matching–fusion mechanism, and robustness to spatial and temporal misalignment.

\textbf{Spatio-Temporal Fusion.}
To validate the complementary contributions of cross-modal and historical evidence, we compare MatchFusion-S and MatchFusion-ST in Table.~\ref{tab:spatiotemporal_ablation}.
MatchFusion-S improves NDS from 70.1 for the LiDAR-only baseline to 72.5. Adding temporal interaction further increases NDS to 73.0, while improving AMOTA from 49.0 to 62.3 and MOTA from 43.6 to 60.8. These results demonstrate that temporal aggregation complements spatial fusion, benefiting broad 3D perception tasks.

The independent applicability of MatchFusion-T is evaluated through its integration into SparseDrive without spatial LiDAR–camera fusion. 
As shown in Table.~\ref{tab:e2e_comparison}, MatchFusion-T improves detection and tracking and \textcolor{black}{reduces planning CR}. 
These results support the utility of MatchFusion for shared instance representations beyond standalone perception.

\textbf{Matching–Fusion Construction.}
Table.~\ref{tab:matching_fusion_ablation} evaluates matchmap construction and feature aggregation within MatchFusion-S.
With direct 3D geometry, structural matching achieves 72.2 NDS and 69.1 mAP. 
Adding semantic refinement further improves detection performance, supporting the contribution of contextual query information to correspondence estimation.
Replacing direct 3D geometry with multi-view image-plane geometry further improves performance to 72.5 NDS and 69.6 mAP. 
These results favor combining projection-based structural priors with semantic refinement for cross-modal matching.
The fusion comparison in Table~\ref{tab:matching_fusion_ablation} evaluates alternative instance aggregation strategies. 
Direct concatenation achieves 69.4 NDS and 64.8 mAP, below the LiDAR-only baseline. 
Self-attention and cross-attention improve NDS to 71.3 and 71.8, respectively, demonstrating the benefit of information exchange between instances. 
MatchFusion-S further improves over cross-attention by 0.7 NDS and 1.7 mAP points, supporting residual aggregation guided by structurally initialized and semantically refined correspondences.
Table~\ref{tab:category_consistency_ablation} evaluates the effect of category consistency. 
Enforcing category consistency improves NDS from 72.1 to 73.0 and mAP from 69.0 to 70.4.
These gains indicate that excluding cross-category associations provides more reliable correspondences for subsequent semantic refinement and feature aggregation. 
% Enforcing category consistency also improves computational efficiency, as shown in Table.~\ref{tab:matchmap_efficiency}.

% \begin{table}[t]
%   \centering
%   \caption{
%     Sensitivity of MatchFusion-ST to the semantic refinement
%     threshold $\epsilon$ on the crowded subset of the nuScenes
%     validation set, containing at least 20 objects per frame.
%   }
%   \label{tab:epsilon_sensitivity}
%   \setlength{\tabcolsep}{12pt}
%   \begin{tabular}{@{}ccc@{}}
%     \toprule
%     $\epsilon$ & NDS$\uparrow$ & mAP$\uparrow$ \\
%     \midrule
%     $10^{1}$   & 72.0 & 69.1 \\
%     $0$        & 72.8 & 69.9 \\
%     $-10^{1}$  & \textbf{72.9} & \textbf{70.0} \\
%     $-10^{2}$  & 72.0 & 69.1 \\
%     $-10^{3}$  & 72.7 & 69.9 \\
%     \bottomrule
%   \end{tabular}
%   \vspace{-3mm}
% \end{table}

\begin{table}[!t]
  \centering
  \caption{
  Effect of front-end optimization on the training efficiency and detection performance of MatchFusion-ST on the nuScenes val set. 
  Training time and GPU memory usage are normalized to the configuration with frozen front-ends.
  }
  \label{tab:frontend_training}
  \setlength{\tabcolsep}{10pt}
  \begin{tabular}{c|cc|cc}
    \toprule
    Front-ends
    & Mem.$\downarrow$ & Time$\downarrow$
    & NDS$\uparrow$ & mAP$\uparrow$ \\
    \midrule
    Unfrozen & 2.1$\times$ & 5.5$\times$ & 71.7 & 68.5 \\
    Frozen & \textbf{1.0$\times$} & \textbf{1.0$\times$}
    & \textbf{73.0} & \textbf{70.4} \\
    \bottomrule
  \end{tabular}
  \vspace{-5mm}
\end{table}

\textbf{Efficiency of Instance-Level Interaction.}
Tables.~\ref{tab:matchmap_efficiency} and~\ref{tab:frontend_training} complement the system-level accuracy–efficiency comparison by evaluating matchmap construction and front-end optimization. 
For MatchFusion-S, structural initialization with selective semantic refinement requires 1.7× training time and 1.3× memory relative to structural-only matching, versus 1.9× and 2.2× for semantic-only matching. 
For perception, MatchFusion-ST with frozen front-ends achieves 73.0 NDS and 70.4 mAP, outperforming the unfrozen configuration while avoiding its 2.1× memory and 5.5× training time requirements. 
These results demonstrate the training efficiency of structural guidance and front-end decoupling, while the jointly trained SparseDrive results in Table I further establish compatibility with end-to-end optimization.

\textbf{Spatio-Temporal Fusion Robustness.}
We evaluate the robustness of MatchFusion-ST by perturbing spatial and temporal priors at inference, as summarized in Table.~\ref{tab:calib_sensitivity}. 
For spatial interaction, translation perturbations of 5–20 cm to the projection extrinsics leave NDS unchanged at the reported precision. 
Rotation perturbations of 1°, 2°, and 4° reduce NDS from 73.0 to 71.9, 71.8, and 71.7, respectively. 
Even at 4°, the system retains 71.7 NDS and 68.0 mAP, exceeding the LiDAR-only baseline by 1.6 and 2.9 points.
These results demonstrate that cross-modal fusion remains beneficial under the tested calibration errors, with greater sensitivity to rotational perturbations.
For temporal interaction, we independently perturb the object motion, ego motion, and inter-frame interval used for historical state alignment. 
As shown in Table.~\ref{tab:calib_sensitivity}, MatchFusion-ST maintains at least 72.3 NDS, 69.2 mAP, 61.2 AMOTA, and 60.1 MOTA across the tested perturbations. 
Detection performance remains above both unimodal baselines, while tracking performance consistently exceeds the camera-only baseline. 
These results demonstrate that the benefits of temporal aggregation persist under imperfect motion compensation.

\begin{table}[t]
  \centering
  \caption{Robustness to spatio-temporal perturbations on MatchFusion-ST.}
  \label{tab:calib_sensitivity}
  \setlength{\tabcolsep}{4pt}
  \begin{tabular}{@{}cc|cc cc@{}}
    \toprule
    \multirow{2}{*}{Perturbation} & \multirow{2}{*}{Magnitude}
    & \multicolumn{2}{c}{Detection}
    & \multicolumn{2}{c}{Tracking} \\
    \cmidrule(lr){3-4}
    \cmidrule(lr){5-6}
    &
    & NDS$\uparrow$ & mAP$\uparrow$
    & AMOTA$\uparrow$ & MOTA$\uparrow$ \\
    \midrule
    \rowcolor{gray!10}
    \multicolumn{6}{c}{\textit{Baseline}} \\
    MatchFusion-ST & ---
      & \textbf{73.0} & \textbf{70.4}
      & \textbf{62.3} & \textbf{60.8} \\
    LiDAR-only detector & ---
      & 70.1 & 65.1
      & --- & --- \\
    Camera-only detector & ---
      & 56.1 & 46.9
      & 49.0 & 43.6 \\
    \midrule
    \rowcolor{gray!10}
    \multicolumn{6}{c}{\textit{Spatial}} \\
    \multirow{3}{*}{Rotation ($^\circ$)}
      & 1 & 71.9 & 68.4 & 59.2 & 56.9 \\
      & 2 & 71.8 & 68.3 & 59.0 & 56.7 \\
      & 4 & 71.7 & 68.0 & 58.8 & 56.5 \\[2pt]
    \multirow{3}{*}{Translation (cm)}
      & 5  & 73.0 & 70.4 & 62.5 & 60.6 \\
      & 10 & 73.0 & 70.4 & 62.5 & 60.4 \\
      & 20 & 73.0 & 70.4 & 62.6 & 60.8 \\
    \midrule
    \rowcolor{gray!10}
    \multicolumn{6}{c}{\textit{Temporal}} \\
    \multirow{3}{*}{$\Delta t$ (s)}
      & 0.1 & 72.9 & 70.3 & 62.4 & 60.2 \\
      & 0.2 & 73.0 & 70.4 & 62.5 & 60.6 \\
      & 0.3 & 72.9 & 70.3 & 62.3 & 60.6 \\[2pt]
    \multirow{3}{*}{Ego Motion (m)}
      & 0.2 & 72.9 & 70.3 & 62.3 & 60.4 \\
      & 0.5 & 72.7 & 69.9 & 62.0 & 60.2 \\
      & 1.0 & 72.3 & 69.2 & 61.3 & 60.3 \\[2pt]
    \multirow{3}{*}{Object Motion (m/s)}
      & 0.5 & 72.8 & 70.2 & 62.1 & 60.6 \\
      & 1.0 & 72.6 & 69.9 & 61.6 & 60.1 \\
      & 2.0 & 72.3 & 69.4 & 61.2 & 60.2 \\
    \bottomrule
  \end{tabular}
\end{table}

\begin{figure}[!t]
  \centering
  \includegraphics[width=0.95\columnwidth]{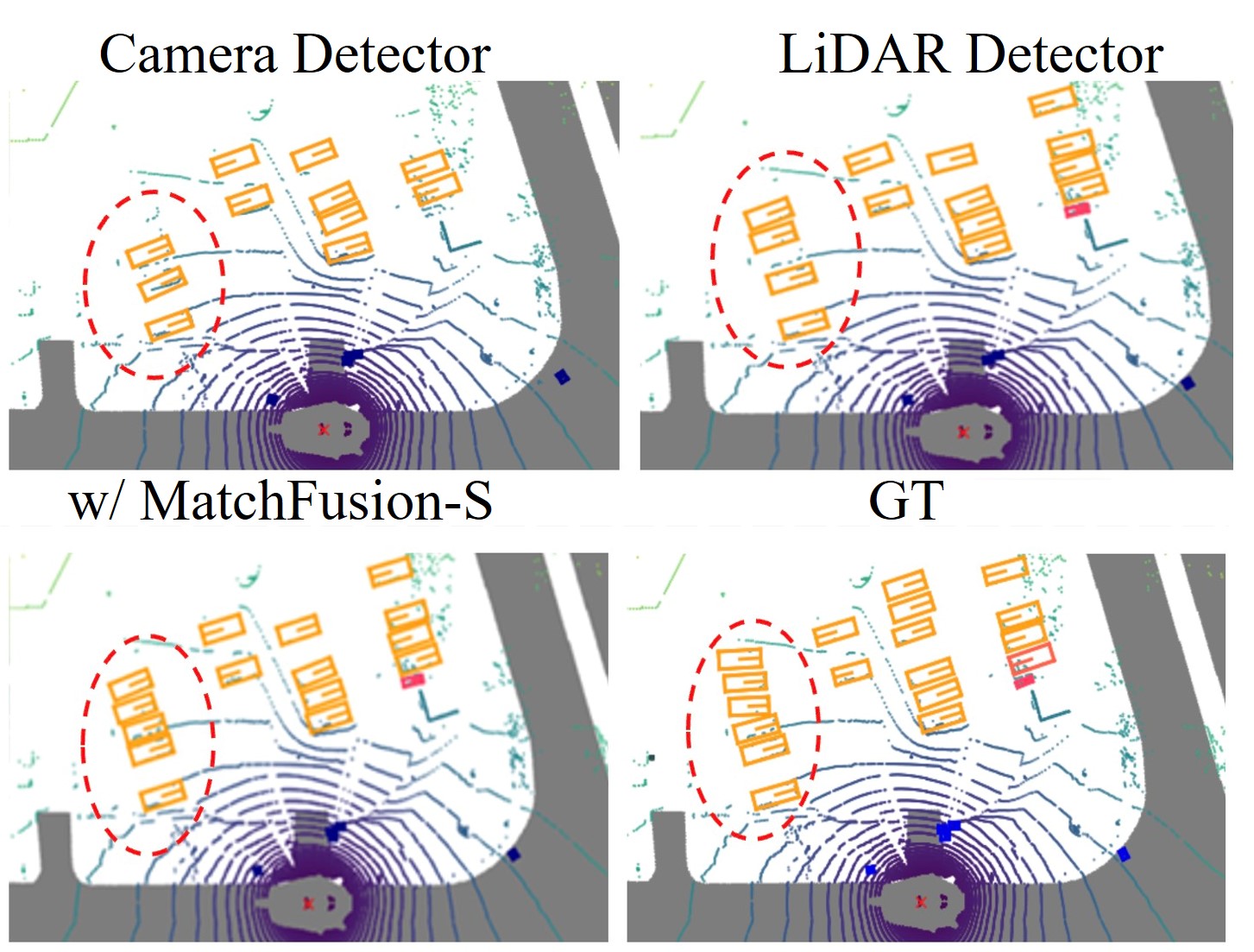}
  \caption{
  Qualitative BEV detection results of MatchFusion-S on the nuScenes validation set.
  Results are shown for the camera front-end Sparse4Dv3, the LiDAR front-end TransFusion-L, and the MatchFusion-equipped system.
  }
  \label{fig:detection_comparison}
\end{figure}

\begin{figure}[!t]
  \centering
  \includegraphics[width=\columnwidth]{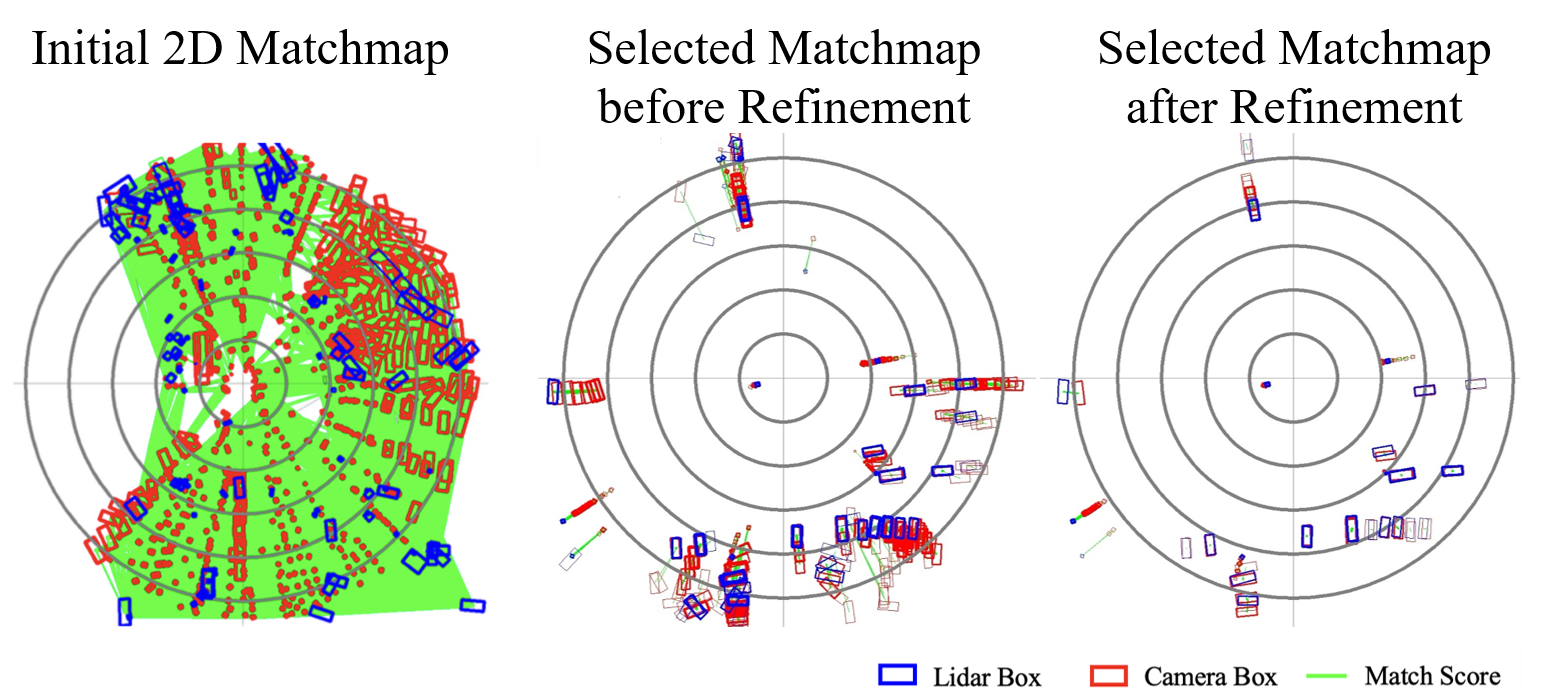}
  \caption{
  Qualitative visualization of explicit--implicit cross-modal matching on the nuScenes validation set.
  Box thickness represents detection confidence, and line thickness represents matching score.
  }
  \label{fig:matching_visualization}
\end{figure}

\subsection{Qualitative Results}
\label{sec:qualitative}

Fig.~\ref{fig:detection_comparison} illustrates the complementary advantages of multimodal fusion.
MatchFusion corrects individual front-end errors and recovers objects missed by a single modality, illustrating the benefit of aggregating complementary evidence through the instance interface.
Fig.~\ref{fig:matching_visualization} highlights the complementary roles of structural initialization and semantic refinement.
Structural priors identify plausible associations, while query semantics distinguish corresponding instances from geometrically ambiguous candidates. 
The refinement gate concentrates semantic corrections on structurally plausible pairs for efficient correspondence refinement.
% In crowded scenes, semantic refinement suppresses spurious associations between geometrically similar instances, as shown in Fig.~\ref{fig:matching_visualization}.
% \textcolor{red}{Fig.~\ref{fig:matching_visualization}(b) illustrates the progression from structural candidates to refined correspondences.
% Compared with direct 3D matching, multi-view image-plane geometry retains candidate associations despite camera depth discrepancies.
% Query semantics then help distinguish compatible pairs from ambiguous candidates.
% These examples illustrate how structural priors and semantic evidence cooperate to produce soft correspondences for residual feature aggregation.

% \begin{figure}[h]
%   \centering
%   % \includegraphics[width=0.95\columnwidth]{Figure/vis.png}
%   \includegraphics[width=\columnwidth]{Figure/vis_new.png}
%   \caption{
%   Qualitative visualization of explicit--implicit cross-modal matching on the nuScenes validation set.
%   Box thickness represents detection confidence, and line thickness represents matching score.
%   }
%   \label{fig:matching_visualization}
% \end{figure}

\FloatBarrier

\section{CONCLUSIONS}
We present MatchFusion, a learnable instance matching and fusion module for spatio-temporal multimodal autonomous driving. 
By combining explicit geometric and categorical priors with selective semantic refinement, MatchFusion establishes soft correspondences for residual feature aggregation.
A common formulation supports spatial LiDAR–camera and temporal past–current interaction while remaining decoupled from front-end feature extraction.
Experiments on nuScenes demonstrate consistent perception gains across diverse front-ends, computational and memory efficiency, and robustness to spatial and temporal perturbations.
Integrating MatchFusion-T into SparseDrive further improves perception under the original training objectives, supporting its compatibility with end-to-end optimization.
These findings support explicit–implicit matching as an efficient and reusable mechanism for instance-level interaction.

\textbf{Future work} will explicitly model uncertainty in geometric priors to adapt their influence and extend MatchFusion to multi-source ensembling for offline perception.

\bibliographystyle{IEEEtran}
\bibliography{IEEEtranBST/IEEEfull}

@inproceedings{caesar2020nuscenes,
  title={nuscenes: A multimodal dataset for autonomous driving},
  author={Caesar, Holger and Bankiti, Varun and Lang, Alex H and Vora, Sourabh and Liong, Venice Erin and Xu, Qiang and Krishnan, Anush and Pan, Yu and Baldan, Giancarlo and Beijbom, Oscar},
  booktitle={CVPR},
  pages={11621--11631},
  year={2020}
}

@inproceedings{li2023poly,
  title={Poly-mot: A polyhedral framework for 3d multi-object tracking},
  author={Li, Xiaoyu and Xie, Tao and Liu, Dedong and Gao, Jinghan and Dai, Kun and Jiang, Zhiqiang and Zhao, Lijun and Wang, Ke},
  booktitle={IROS},
  pages={9391--9398},
  year={2023},
  organization={IEEE}
}

@inproceedings{kim2021eagermot,
  title={Eagermot: 3d multi-object tracking via sensor fusion},
  author={Kim, Aleksandr and O{\v{s}}ep, Aljo{\v{s}}a and Leal-Taix{\'e}, Laura},
  booktitle={ICRA},
  pages={11315--11321},
  year={2021},
  organization={IEEE}
}

@misc{zhang2023bytetrackv2,
      title={ByteTrackV2: 2D and 3D Multi-Object Tracking by Associating Every Detection Box}, 
      author={Yifu Zhang and Xinggang Wang and Xiaoqing Ye and Wei Zhang and Jincheng Lu and Xiao Tan and Errui Ding and Peize Sun and Jingdong Wang},
      year={2023},
      eprint={2303.15334},
      archivePrefix={arXiv},
      primaryClass={cs.CV}
}

@inproceedings{liu2023bevfusion,
  title={Bevfusion: Multi-task multi-sensor fusion with unified bird's-eye view representation},
  author={Liu, Zhijian and Tang, Haotian and Amini, Alexander and Yang, Xinyu and Mao, Huizi and Rus, Daniela L and Han, Song},
  booktitle={ICRA},
  pages={2774--2781},
  year={2023},
  organization={IEEE}
}

@inproceedings{DETR3D,
  title={Detr3d: 3d object detection from multi-view images via 3d-to-2d queries},
  author={Wang, Yue and Guizilini, Vitor Campagnolo and Zhang, Tianyuan and Wang, Yilun and Zhao, Hang and Solomon, Justin},
  booktitle={CoRL},
  pages={180--191},
  year={2022},
  organization={PMLR}
}

@article{lin2023sparse4d,
  title={Sparse4d v3: Advancing end-to-end 3d detection and tracking},
  author={Lin, Xuewu and Pei, Zixiang and Lin, Tianwei and Huang, Lichao and Su, Zhizhong},
  journal={arXiv preprint arXiv:2311.11722},
  year={2023}
}

@inproceedings{tang2025simpb,
  title={Simpb: A single model for 2d and 3d object detection from multiple cameras},
  author={Tang, Yingqi and Meng, Zhaotie and Chen, Guoliang and Cheng, Erkang},
  booktitle={ECCV},
  pages={1--17},
  year={2024},
  organization={Springer}
}

@inproceedings{philion2020lift,
  title={Lift, splat, shoot: Encoding images from arbitrary camera rigs by implicitly unprojecting to 3d},
  author={Philion, Jonah and Fidler, Sanja},
  booktitle={ECCV},
  pages={194--210},
  year={2020},
  organization={Springer}
}

@inproceedings{wang2023exploring,
  title={Exploring object-centric temporal modeling for efficient multi-view 3d object detection},
  author={Wang, Shihao and Liu, Yingfei and Wang, Tiancai and Li, Ying and Zhang, Xiangyu},
  booktitle={ICCV},
  pages={3621--3631},
  year={2023}
}

@inproceedings{bai2022transfusion,
  title={Transfusion: Robust lidar-camera fusion for 3d object detection with transformers},
  author={Bai, Xuyang and Hu, Zeyu and Zhu, Xinge and Huang, Qingqiu and Chen, Yilun and Fu, Hongbo and Tai, Chiew-Lan},
  booktitle={CVPR},
  pages={1090--1099},
  year={2022}
}

@inproceedings{zhou2018voxelnet,
  title={Voxelnet: End-to-end learning for point cloud based 3d object detection},
  author={Zhou, Yin and Tuzel, Oncel},
  booktitle={CVPR},
  pages={4490--4499},
  year={2018}
}

@inproceedings{vora2020pointpainting,
  title={Pointpainting: Sequential fusion for 3d object detection},
  author={Vora, Sourabh and Lang, Alex H and Helou, Bassam and Beijbom, Oscar},
  booktitle={CVPR},
  pages={4604--4612},
  year={2020}
}

@inproceedings{wang2021pointaugmenting,
  title={Pointaugmenting: Cross-modal augmentation for 3d object detection},
  author={Wang, Chunwei and Ma, Chao and Zhu, Ming and Yang, Xiaokang},
  booktitle={CVPR},
  pages={11794--11803},
  year={2021}
}

@article{yang2022deepinteraction,
  title={Deepinteraction: 3d object detection via modality interaction},
  author={Yang, Zeyu and Chen, Jiaqi and Miao, Zhenwei and Li, Wei and Zhu, Xiatian and Zhang, Li},
  journal={NeurIPS},
  volume={35},
  pages={1992--2005},
  year={2022}
}

@inproceedings{xie2023sparsefusion,
  title={Sparsefusion: Fusing multi-modal sparse representations for multi-sensor 3d object detection},
  author={Xie, Yichen and Xu, Chenfeng and Rakotosaona, Marie-Julie and Rim, Patrick and Tombari, Federico and Keutzer, Kurt and Tomizuka, Masayoshi and Zhan, Wei},
  booktitle={ICCV},
  pages={17591--17602},
  year={2023}
}

@article{yang2025deepinteraction++,
  title={Deepinteraction++: Multi-modality interaction for autonomous driving},
  author={Yang, Zeyu and Song, Nan and Li, Wei and Zhu, Xiatian and Zhang, Li and Torr, Philip HS},
  journal={TPAMI},
  year={2025},
  publisher={IEEE}
}

@inproceedings{liao2025diffusiondrive,
  title={Diffusiondrive: Truncated diffusion model for end-to-end autonomous driving},
  author={Liao, Bencheng and Chen, Shaoyu and Yin, Haoran and Jiang, Bo and Wang, Cheng and Yan, Sixu and Zhang, Xinbang and Li, Xiangyu and Zhang, Ying and Zhang, Qian and others},
  booktitle={CVPR},
  pages={12037--12047},
  year={2025},
  organization={IEEE}
}

@inproceedings{sun2025sparsedrive,
  title={Sparsedrive: End-to-end autonomous driving via sparse scene representation},
  author={Sun, Wenchao and Lin, Xuewu and Shi, Yining and Zhang, Chuang and Wu, Haoran and Zheng, Sifa},
  booktitle={ICRA},
  pages={8795--8801},
  year={2025},
  organization={IEEE}
}

@article{chitta2022transfuser,
  title={Transfuser: Imitation with transformer-based sensor fusion for autonomous driving},
  author={Chitta, Kashyap and Prakash, Aditya and Jaeger, Bernhard and Yu, Zehao and Renz, Katrin and Geiger, Andreas},
  journal={TPAMI},
  volume={45},
  number={11},
  pages={12878--12895},
  year={2022},
  publisher={IEEE}
}

@inproceedings{song2025momad,
  title={Don’t shake the wheel: Momentum-aware planning in end-to-end autonomous driving},
  author={Song, Ziying and Jia, Caiyan and Liu, Lin and Pan, Hongyu and Zhang, Yongchang and Wang, Junming and Zhang, Xingyu and Xu, Shaoqing and Yang, Lei and Luo, Yadan},
  booktitle={CVPR},
  pages={22432--22441},
  year={2025},
  organization={IEEE}
}

@article{wang2025mv2dfusion,
  title={Mv2dfusion: Leveraging modality-specific object semantics for multi-modal 3d detection},
  author={Wang, Zitian and Huang, Zehao and Gao, Yulu and Wang, Naiyan and Liu, Si},
  journal={TPAMI},
  year={2025},
  publisher={IEEE}
}

@ARTICLE{StreamCMT,
  author={Huang, Yanliang and Liu, Yuansheng},
  journal={RA-L}, 
  title={StreamCMT: Prior-Guided Multimodal Temporal Fusion for Sparse 3D Object Detection}, 
  year={2026},
  volume={11},
  number={5},
  pages={5358-5365},
  doi={10.1109/LRA.2026.3671538}}

@inproceedings{li2026rethinking,
  title={Rethinking the spatio-temporal alignment of end-to-end 3D perception},
  author={Li, Xiaoyu and Li, Peidong and Wu, Xian and Shi, Long and Liu, Dedong and Wu, Yitao and Fu, Jiajia and Cui, Dixiao and Zhao, Lijun and Sun, Lining},
  booktitle={AAAI},
  volume={40},
  number={8},
  pages={6513--6520},
  year={2026}
}

@article{wu2026fusion,
  title={Fusion-Poly: A Polyhedral Framework Based on Spatial-Temporal Fusion for 3D Multi-Object Tracking},
  author={Wu, Xian and Wu, Yitao and Li, Xiaoyu and Li, Zijia and Zhao, Lijun and Sun, Lining},
  journal={arXiv preprint arXiv:2603.08199},
  year={2026}
}

@inproceedings{nguyen2026lead,
  title={Lead: Minimizing learner-expert asymmetry in end-to-end driving},
  author={Nguyen, Long and Fauth, Micha and Jaeger, Bernhard and Dauner, Daniel and Igl, Maximilian and Geiger, Andreas and Chitta, Kashyap},
  booktitle={CVPR},
  pages={39775--39785},
  year={2026}
}

@inproceedings{zhang2025bridging,
  title={Bridging past and future: End-to-end autonomous driving with historical prediction and planning},
  author={Zhang, Bozhou and Song, Nan and Jin, Xin and Zhang, Li},
  booktitle={CVPR},
  pages={6854--6863},
  year={2025},
  organization={IEEE}
}

@inproceedings{he2016deep,
  title={Deep residual learning for image recognition},
  author={He, Kaiming and Zhang, Xiangyu and Ren, Shaoqing and Sun, Jian},
  booktitle={CVPR},
  pages={770--778},
  year={2016}
}

@article{adamw,
  title={Decoupled weight decay regularization},
  author={Loshchilov, Ilya and Hutter, Frank},
  journal={arXiv preprint arXiv:1711.05101},
  year={2017}
}

\end{document}